\pdfoutput=1

\documentclass[11pt]{article}

\usepackage{acl}

\usepackage{times}
\usepackage{latexsym}

\usepackage[T1]{fontenc}

\usepackage[utf8]{inputenc}

\usepackage{microtype}

\usepackage{amsmath,graphicx,amssymb,bm,subfigure}
\usepackage{booktabs}
\usepackage{multirow}
\usepackage{makecell}
\usepackage{color}
\title{Event Causality Extraction with Event Argument Correlations}

\author{Shiyao Cui$^{1,2}$ \ Jiawei Sheng$^{1,2}$ \ Xin Cong$^{1,2}$ \\ \textbf{QuanGang Li}$^{1,2}$\thanks{\hspace{0.15cm}Corresponding   Author} \ \textbf{Tingwen Liu}$^{1,2}$ \ \textbf{Jinqiao Shi}$^{3,1}$  \\
	$^1$Institute of Information Engineering, Chinese Academy of Sciences. Beijing, China \\
	$^2$School of Cyber Security, University of Chinese Academy of Sciences. Beijing, China \\
	$^3$Beijing University of Posts and Telecommunications. Beijing, China \\
	{\tt \{cuishiyao, shengjiawei, congxin, liquangang\}@iie.ac.cn} \\
	{\tt liutingwen@iie.ac.cn  shijinqiao@bupt.edu.cn}\\
}

\begin{document}
\maketitle

\begin{abstract}
Event Causality Identification (ECI), which aims to detect whether a causality relation exists between two given textual events, is an important task for event causality understanding.
However, the ECI task ignores crucial event structure and cause-effect causality component information, making it struggle for downstream applications.
In this paper, we explore a novel task, namely Event Causality Extraction (ECE), aiming to extract the cause-effect  event causality pairs with their structured event information from plain texts.
The ECE task is more challenging since each event can contain multiple event arguments, posing fine-grained correlations between events to decide the cause-effect event pair.
Hence, we propose a method with a dual grid tagging scheme to capture the intra- and  inter-event argument correlations for ECE.
Further, we devise a event type-enhanced model architecture to realize the dual grid tagging scheme.
Experiments demonstrate the effectiveness of our method, and extensive analyses point out several future directions for ECE.
\end{abstract}

\section{Introduction}
\label{sec:introduction}

\begin{figure}[t]
	\centering
	\includegraphics[width=1.0\linewidth]{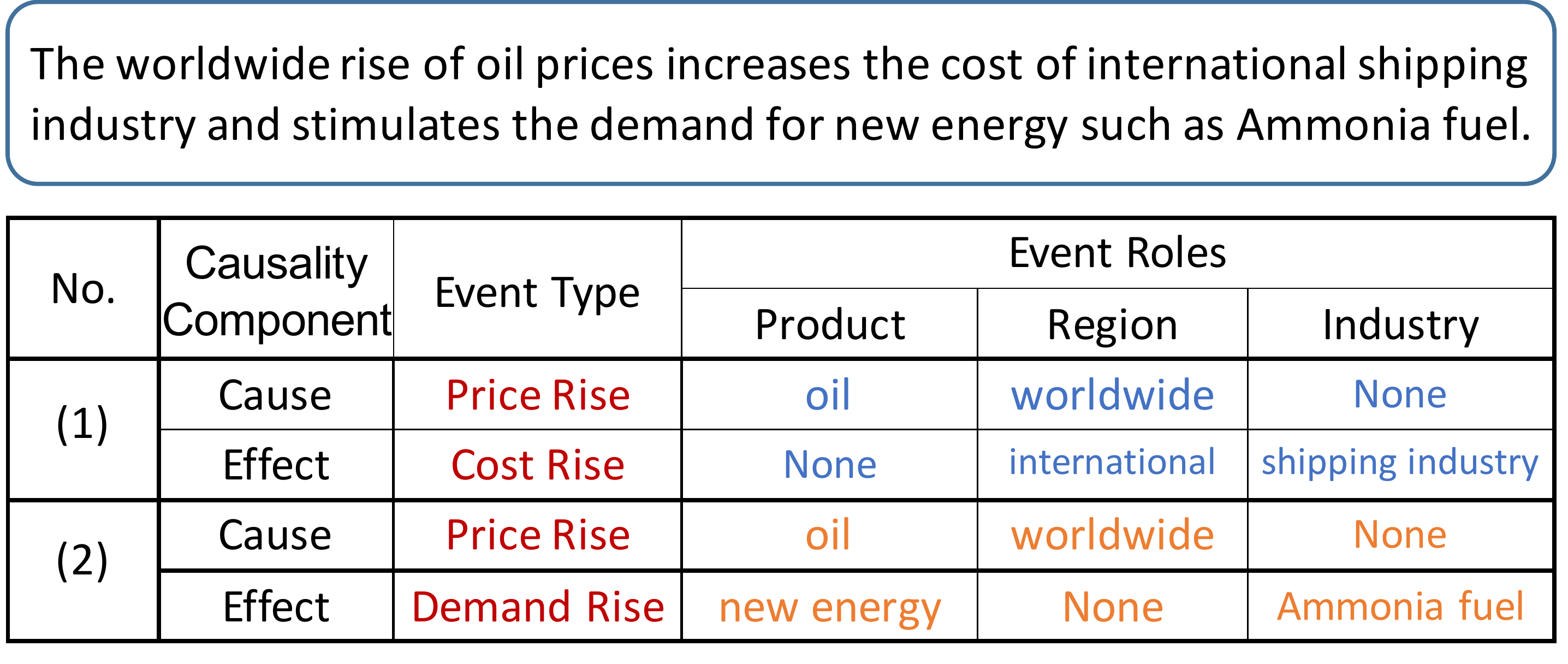}
	\caption{Illustration for ECE which takes the raw text as input, and outputs the structured event causality pair.}
	\label{fig:ece}
\end{figure}

Event causality~\cite{liu2020KnowledgeEE,cao-etal-2021-knowledge} denotes an explicit causal relation between two events, constituting a specific
cause-effect event pair.
As shown in Figure~\ref{fig:ece}, a causal relation exists between a \texttt{Price Rise} event (\textit{The worldwide rise of oil prices}) and a \texttt{Cost Rise} event (\textit{increases the cost of international shipping industry}).
Understanding such event causality could facilitate various downstream applications including event forecasting~\cite{hashimoto-etal-2014-toward}, intelligent search~\cite{Rudnik2019SearchingNA} and question answering~\cite{Costa2020EventQAAD}, which is important for natural language understanding.

In recent years, it has aroused the research interest for Event Causality Identification (ECI)~\cite{liu2020KnowledgeEE,cao-etal-2021-knowledge,zuo-etal-2021-improving,zuo-etal-2021-learnda,zuo-etal-2020-knowdis,tran-phu-nguyen-2021-graph}, which aims to detect whether the causality exists between two given events.
Despite of its success, there exist two issues that the ECI task fails to address.
1) \textbf{Event Structure Missing}, where each event in ECI is only expressed using a word or phrase which reflects its occurrence, but ignores the explicit event type and event arguments (i.e., entities which participate in the event).
%
The absence of such event structure would lose valuable clues for understanding event causality.
As shown in Figure~\ref{fig:ece}, ``oil'' plays a \texttt{Product} role in a \texttt{Price Rise}-typed cause event, implying a consequent \texttt{Cost Rise}-typed effect event towards ``shipping industry''.
2) \textbf{Causality Component Missing}, where ECI only predicts the existence of causality between the given event pairs, ignoring to discriminate the specific cause/effect event causality component.
Limited by these issues, ECI insufficiently explores the causality between events, which demands further promotion to the understanding of event causality. 

Motivated by discussion about event causality in ~\citeauthor{CCKS2021EventCausal}~\shortcite{CCKS2021EventCausal}, we therefore formulate a task termed as \textbf{Event Causality Extraction (ECE)}. As Figure~\ref{fig:ece} shows, ECE aims to end-to-end extract the cause-effect event pairs with structured event information from plain texts.
Comparing with ECI, ECE illustrates the event causality including the event structure, namely event types and arguments, and the specific cause-effect causality component, making it more informative to support the various downstream applications~\cite{DBLP:conf/www/WangYGSMCZL21}.

Intuitively, ECE could be achieved by successively extracting the structured event and then classifying their causality relation.
Unfortunately, such a paradigm would easily suffer from the redundant event-pair problem, where the causality-unrelated events would be inevitably extracted, confusing the causality decision.
Another promising direction is to borrow ideas from relational triple extraction (RTE), which shares the similar task formulation.
However, comparing with the entity-centric RTE task, the event-centric ECE raises new challenges:
1) \textbf{Intra-event Argument Correlations}.
Specifically, ECE focuses on event, which is a structure maintaining interactive correlations among its arguments.
For example in Figure~\ref{fig:ece}, the argument ``new energy'' and ``Ammonia fuel'' in the \texttt{Demand Rise} event have strong semantic correlation.
While RTE focuses on individual entity, thus simply adapting RTE models cannot capture such correlations to derive the event structure.
2) \textbf{Inter-event Argument Correlations}. 
Concretely, the event arguments involved in a cause/effect event pair usually show semantic correlations for causality deduction.
As shown in Figure~\ref{fig:ece}, event \texttt{Pricing\_Rise} which occurs in ``worldwide'' \texttt{Region} could imply event \texttt{Cost\_Rise} in ``international'' \texttt{Region}.
It demonstrates that the inter-event argument correlations not only provide important clues to decide causality, but also benefit reliable cause/effect event extraction with mutual confirmation between the cause-effect pair.

In this paper, we propose an effective approach named \textbf{DualCor}, which explores both the intra-event and inter-event argument \underline{Cor}relations with a \underline{dual} grid tagging scheme for ECE.
Specifically, DualCor contains two grid tagging tables regarding event types and the input sentence, to respectively derive the event structures for cause and effect events. 
In each table, DualCor extracts structured event arguments according to different event types, naturally considering intra-event argument correlations.
Further, when predicting the event arguments in the cause/effect table, DualCor also predicts their corresponding effect/cause event arguments, serving as auxiliary arguments to promote inter-event argument correlations.
By confirming the auxiliary arguments in the other table, DualCor matches reliable cause-effect event pairs as predictions.
To realize the above dual grid tagging scheme, we further devise a type-aware encoder, which refines textual representations with essential event type information to enhance argument prediction. 
We conduct the dual grid tagging on the type-aware textual representations to derive the final cause-effect event pair.
Overall, our main contributions include:

(1) To promote the understanding to event causality, we formulate a new task named  Event Causality Extraction (ECE), which succeeds ECI to push forward the research of event causality understanding. 

(2) We propose a novel approach, DualCor, to exploit the intra-event and inter-event argument correlations for ECE, and present it as a baseline to inspire the following research.

(3) Experiments\footnote{Dataset and source code for implementation are available here: https://github.com/cuishiyao96/ECE} on the ECE dataset reflect the effectiveness of DualCor, and extensive analyses show potential research directions for future works.

\section{Related Works}

This paper explores a novel ECE task, which aims to extract the cause-effect event pairs with structured event information from plain texts.
Existing event-causality-related researches mostly focus on event causality identification, which predicts the causality for the previously given event pairs. 
They can be roughly categorized into three groups:
(1) Early works exploit the linguistic features~\cite{riaz-girju-2013-toward, gao-etal-2019-modeling},  causal patterns~\cite{hu-etal-2017-inference,do-etal-2011-minimally} and statistical causal associations~\cite{riaz-girju-2014-depth} to explore the causality between events.
(2) Recent researchers~\cite{liu2020KnowledgeEE,cao-etal-2021-knowledge,zuo-etal-2021-improving,zuo-etal-2021-learnda,zuo-etal-2020-knowdis} pay the major focus on incorporating external knowledge for causality identification with limited training data.
(3) Different from works above which conduct ECI within  a single sentence,   \citeauthor{tran-phu-nguyen-2021-graph}~\shortcite{tran-phu-nguyen-2021-graph} focus on document-level ECI, where the given events scatter in multiple sentences.
Despite of their success, they all suffer from the issues of \textbf{event structure missing} and \textbf{causality component missing}.
To our best knowledge,  ECE is the first to  simultaneously derive the structured event information and explicit causality component, which could better support the downstream applications. 

Other than ECE, Relational Triple Extraction (RTE)~\cite{DBLP:conf/ecai/0002ZSLWWL20,10.1145/3477495.3531831} has the similar task formulation, the ideas of which could actually be adapted for ECE.
Concretely, RTE detects entity pairs in a sentence and predicts pre-defined relation types between them.
Existing approaches for RTE could be roughly categorized into two lines.
(1) Traditional joint methods, which solve RTE through sequential interrelated steps via task decomposition~\cite{wei-etal-2020-novel, DBLP:conf/ecai/0002ZSLWWL20, DBLP:conf/pkdd/Cong0LCT020} or sequence generation~\cite{zeng-etal-2018-extracting, DBLP:conf/aaai/NayakN20}. 
Unfortunately, these methods all suffer from the \textbf{exposure bias}~\cite{wang-etal-2020-tplinker} problem due to the  gap from training to inference between multiple steps. 
(2) Unified joint methods, which simultaneously derive the triplet entities and relations in one-stage without cascading between steps, and is thus free from the exposure bias.
These methods solve RTE in either a sequence-labeling manner~\cite{zheng-etal-2017-joint} or grid-filling manner~\cite{wang-etal-2021-unire, wang-etal-2020-tplinker}.
However, the entity-centric RTE methods seem to struggle for the event-centric task, since events present more complicated argument correlations either intra- and inter events.

\section{Task Formulation}

Event causality extraction~(ECE) aims to derive the cause-effect event pairs from plain texts.
Here, a cause-effect event pair contains a \textit{Cause component} and an \textit{Effect component}, where each component denotes an event with a specific \textit{event type} and its \textit{event arguments} with their \textit{event roles}.
Given a piece of text, an event causality extraction system is required to predict all the cause-effect event pairs from it as Figure~\ref{fig:ece} shows.

\section{Dual Grid Tagging Scheme}
\begin{figure}[t]
	\centering
	\subfigure[Grid tagging for in the cause table.]{
			\includegraphics[width=1.0\linewidth]{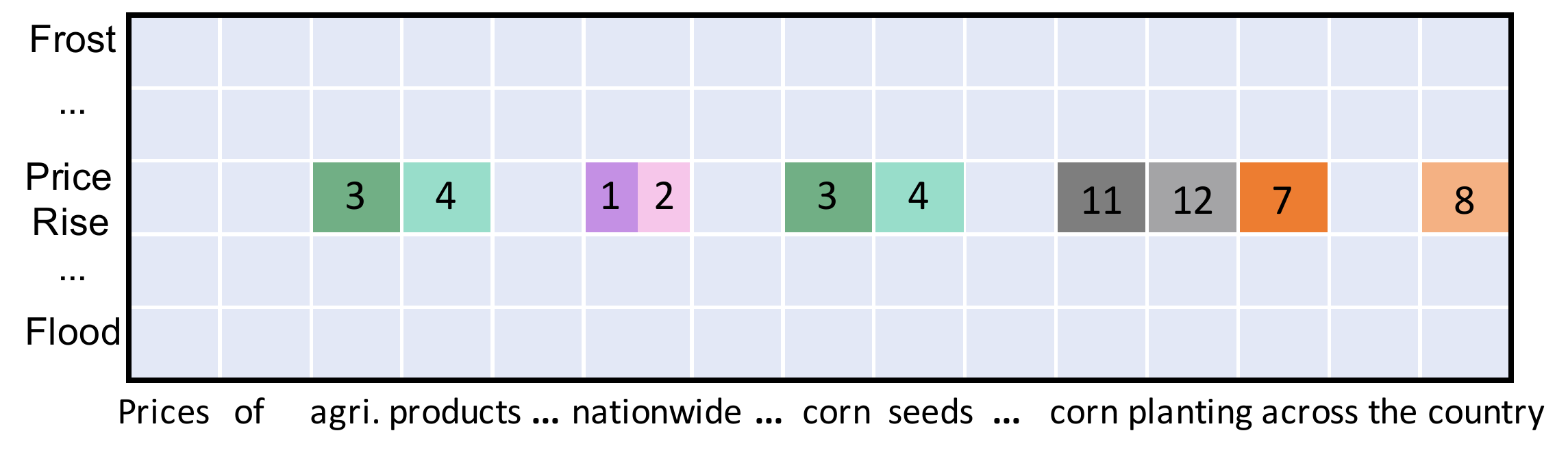}
			\label{fig:cause}
	}
	\subfigure[Grid tagging in the effect table.]{
			\includegraphics[width=1.0\linewidth]{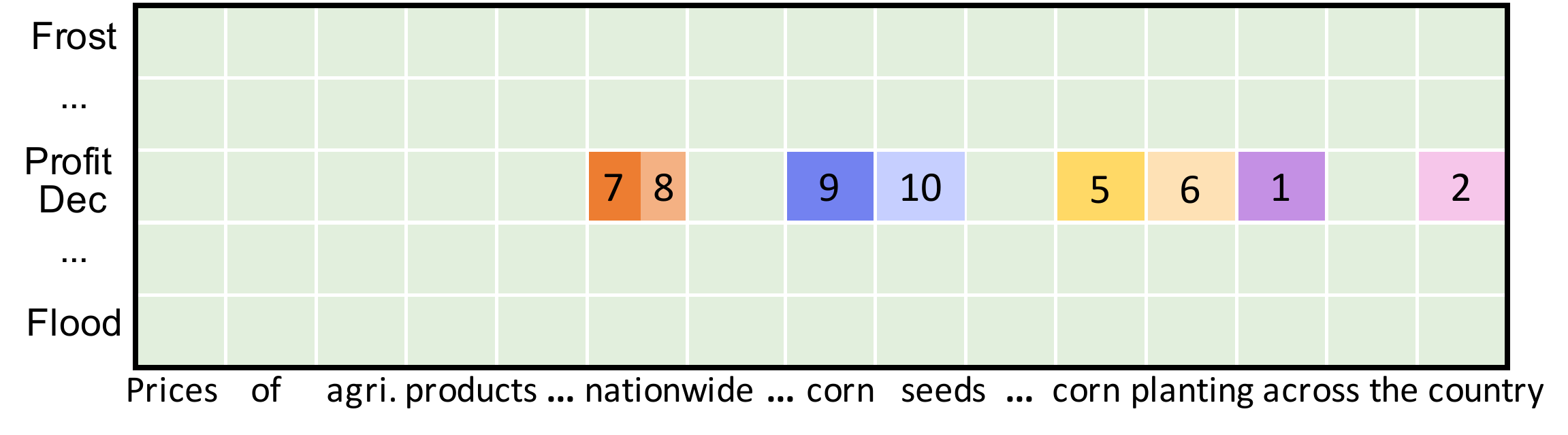}
			\label{fig:effect}
	}
	\subfigure[Tags map.]{
			\includegraphics[width=1.0\linewidth]{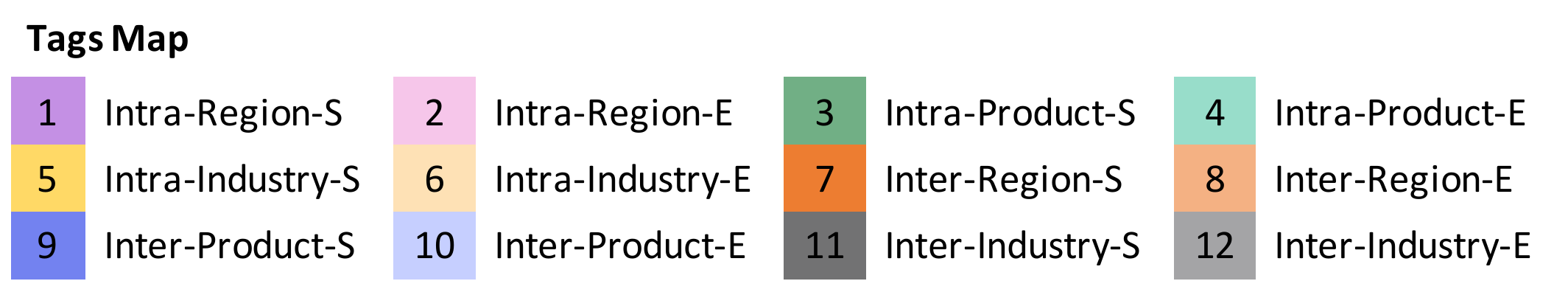}
			\label{fig:tags}
	}
	\caption{Tagging scheme illustration with the sentence ``Prices of agricultural products rose, but the nationwide soaring prices of corn seeds decreased the profit in corn planting across the country.'', where the boundary-field is short as ``S'' and ``E''.} \label{fig:dualgridtagging}
\end{figure}

This section introduces our proposed dual grid tagging scheme for the ECE, including the tagging scheme and its decoding strategy.
The specific model implementation is introduced in Section.~\ref{sec:model}.

\subsection{Tagging Scheme}
\label{sec:tagging_scheme}
In general, we construct two grid tagging tables respectively for the cause/effect events, where each table extracts all the possible events occurring in the sentence.
Formally, given an $n$-token sentence and $m$ predefined event types, we construct two $m \times n $ grid tables for the cause and effect events respectively.
As shown in Figure~\ref{fig:dualgridtagging}, each row denotes arguments within the same event type, while each column denotes tags assigned to the token in the sentence based on the event type.
For each entry in the tables, we fill it with a tag in the form of \begin{math} \{ \mathtt{Cor\text{-}Rol\text{-}Bdy} \} \end{math} consisting of three fields, namely \textbf{correlation-field}, \textbf{role-field} and \textbf{boundary-field}:

(1) For the boundary-field:  $\mathtt{Bdy} \in \{\mathtt{Sta}, \mathtt{End}\}$, we devise it to denote start and end position of argument spans.
For example in Figure~\ref{fig:cause}, we match the argument ``corn seeds'' by matching the $\mathtt{Cor\text{-}Rol\text{-}Sta}$ and $\mathtt{Cor\text{-}Rol\text{-}End}$ tags.

(2) For the role-field: $\mathtt{Rol}\!\!\in\!\!\{\mathtt{Rol}_i\}_{i}$ ($i$ for role index), we devise it to denote the event role for each argument in an event, thus constituting an event structure.
For example in Figure~\ref{fig:cause}, we decide the argument ``corn seeds'' as a \texttt{Product}-role argument in \texttt{Price\_Rising}-type event based on its $\mathtt{Cor\text{-}Product\text{-}Bdy}$ tag in \texttt{Price\_Rising} row.

(3) For correlation-field: $\mathtt{Cor}\!\!\in\!\!\{\mathtt{Intra},\mathtt{Inter}\}$, we devise it to denote event argument correlations in the cause-effect event pair.
Specifically, $\mathtt{Intra}$ denotes the arguments belonging to the same event in the one causality component, while $\mathtt{Inter}$ denotes the arguments in the other causality component.
For example, when predicting the cause event in the cause table, we predict not only the cause arguments (marked with $\mathtt{Intra}$) with cause event type, but also the potential effect arguments (marked with $\mathtt{Inter}$) as auxiliary arguments for mutual confirmation in causality pair matching.
As Figure~\ref{fig:cause} shows, we not only predict argument ``corn seeds'' with $\mathtt{Intra}$ for the \texttt{Price\_Rise}-type cause event, but also predict ``corn planting'' with $\mathtt{Inter}$ tag as effect event argument. 
By matching argument ``corn planting'' with $\mathtt{Intra}$ tag in the effect table, we can derive a \texttt{Price\_Rise}-typed and \texttt{Profit\_Declination}-typed event pair.

Building upon the tagging scheme, the model can naturally extract causality event pairs with their arguments.
Besides, the scheme learns event arguments for each type within separate type row, allowing the model to consider intra-event argument correlations with type-specific information.
Moreover, the tagging scheme enforces the model to extract arguments in one causality component perceiving arguments in the other causality component, thus capturing inter-event argument correlations.

\subsection{Decoding Strategy}

Based upon the tagging scheme, we introduce the decoding strategy for the tagging results. 
Specifically, we decompose the process into three steps, including argument span decoding, event structure decoding and causality pair decoding. Appendix~\ref{appendix:decoding} also provides figure illustration to these three steps. 
\textbf{Step 1. Argument span decoding.}
To derive argument spans for cause/effect events, we adopt the nearest start-end match principle~\cite{wei-etal-2020-novel}. 
Specifically, for entry tags having the same correlation-field and role-field in the same row, we match the start position to the nearest end position according to the position-field to obtain candidate argument spans.
For example in Figure~\ref{fig:cause}, this step ought to predict ``agriculture products'', ``nationwide'', ``corn seeds'', ``corn planting'' and ``across the country'' as candidate argument spans.

\textbf{Step 2. Event structure decoding.}
To derive event structure for cause/effect events, we collect candidate argument spans attached to the same event type.
Specifically, we merge the event arguments with correlation-field $ \mathtt{Intra}$ belonging to the same row, resulting in structured candidate events.
For example in Figure~\ref{fig:cause}, given the candidate argument spans in Step 1, this step ought to select ``agriculture products'', ``nationwide'' and ``corn seeds'' with $\mathtt{Intra}$ tags as the candidate  \texttt{Price\_Rising}-type cause event arguments.

\textbf{Step 3. Causality pair decoding.}
To derive causality pairs, we match inter-event correlated arguments between candidate cause and effect events. 
Specifically, we search the arguments co-occurring in both event tables simultaneously associating with correlation-field $\mathtt{Intra}$ and $\mathtt{Inter}$, and then confirm cause-effect event arguments.
For example in Figure~\ref{fig:cause}, given the candidate event arguments in Step 2, this step ought to select ``nationwide'' and ``corn seeds'' as the true cause event arguments, since there also exist ``nationwide'' and ``corn seeds'' with $\mathtt{Inter}$ tags in the effect tables (Figure~\ref{fig:effect}).
Similarly, this step also selects ``corn planting'', ``across the country'' as the arguments in the \texttt{Profit\_Declination}-type effect events.
Accordingly, it predicts the \texttt{Price\_Rise}-type cause and \texttt{Profit\_Declination}-type effect event pair as  Figure~\ref{fig:model} shows. 
Note that  though  ``agriculture product'' is also an event argument candidate of a \texttt{Price\_Rise}-type event in Step 2, it is not  included in the causality pair due to the absence of $\mathtt{Inter}$ correlation in the effect table.

\section{Model}~\label{sec:model}
\begin{figure}[t]
	\centering
	\includegraphics[width=0.95\linewidth]{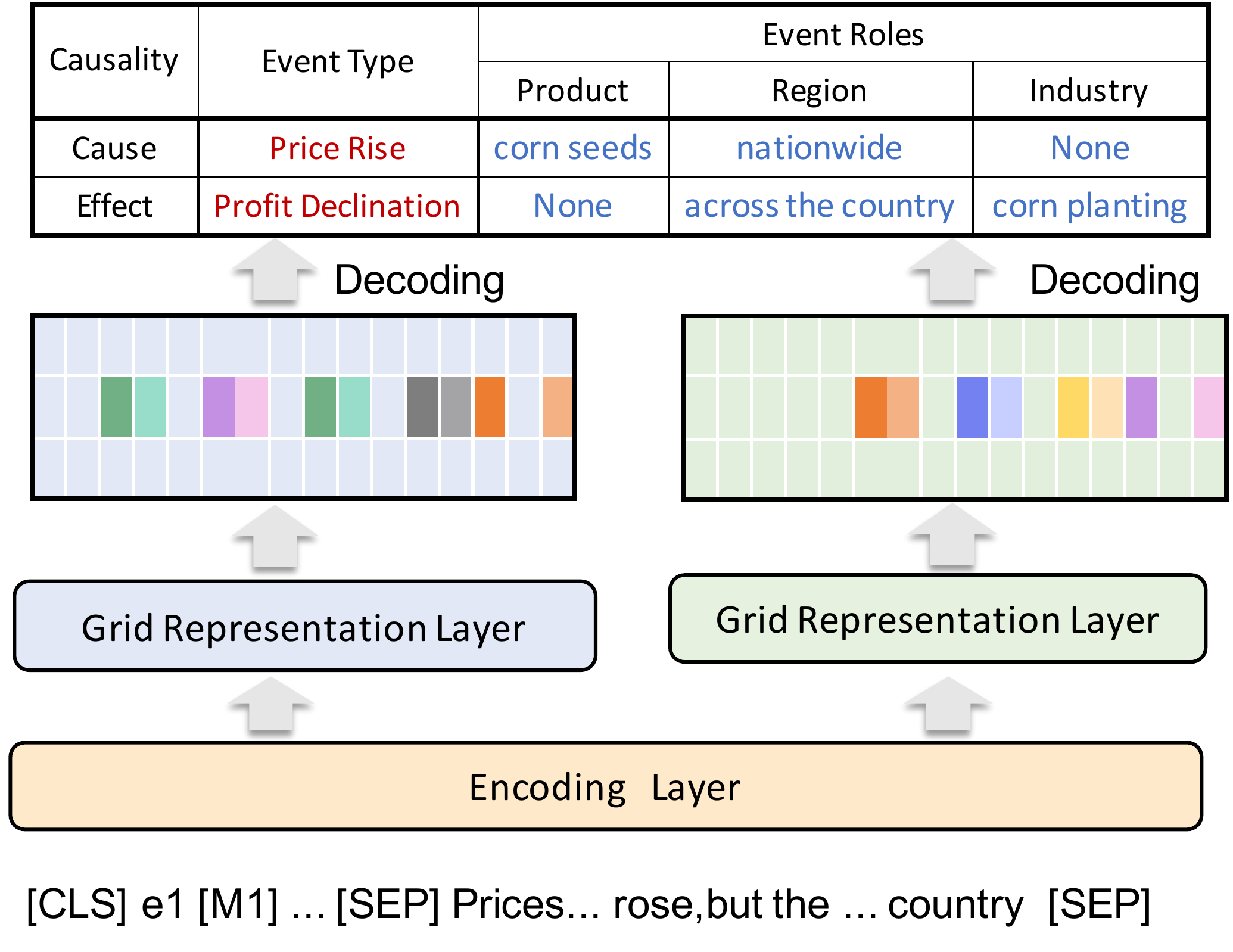}
	\caption{A toy illustration to our model architecture.}
	\label{fig:model}
\end{figure}

In this section, we introduce the model architecture to implement DualCor as Figure~\ref{fig:model} shows.

\subsection{Encoding Layer}
\label{sec:encoding}

This layer derives the contextualized representations of tokens in the sentence and event types.
To facilitate the following event argument prediction, we intend to conduct \textbf{event type-aware encoding} which refines textual representations with event type information.
Specifically, we concatenate the event types ahead of the sentence, and employ BERT~\cite{devlin-etal-2019-bert} for encoding thanks to its deep self-attention architectures~\cite{NIPS2017_3f5ee243}.
Supposing that a text consisting of $n$ tokens $\{t_1, t_2, ..., t_n\}$ and $m$ predefined  event types $\{e_1, e_2, ..., e_m\}$ are given,   the input sequence is organized  in the form as follows: 
\begin{equation}
\label{equ:encoding}
\mathtt{[CLS]~e_1~[M_1]~e_2~[M_2]... ~e_m~[M_{m}]~[SEP]} \mathtt{t_1 ~ ... ~ t_n ~ [SEP]}
\end{equation}
where  $\mathtt{[M_j]}$ is the marker for the $j_{th}$ event types $e_j$.
We feed the input sequence into the encoder and use the output representations $ \mathbf{H} = \mathbf{h}_1, \mathbf{h}_2, ..., \mathbf{h}_n$ corresponding to the sentence as token representations.
Then, we gather representations  of event type markers as event type representations, which is denoted as $\mathbf{E} = \mathbf{e}_1, \mathbf{e}_2, ..., \mathbf{e}_m$.

\subsection{Grid Representation Layer}

This section first details the function for producing entry representations, and then introduces how to apply it in both grid tables.

\subsubsection{Semantic Fusion Function}
\label{sec:fusion}
Each entry in the grid respectively models the relation between one token and  an event type for event argument deduction.
For an entry connecting the $j_{\rm{th}}$ event type $e_j$ and $i_{\rm{th}}$ token in the sentence,  its representation $\mathbf{g}_{j,i}$ could be obtained  via a fusion function $\phi$ by integrating the semantics of $t_i$ and $e_j$ as $\mathbf{g}_{j,i} = \phi(\mathbf{e}_j, \mathbf{h}_i)$.
Intuitively, $\phi$ could be achieved in various semantic fusion ways including concatenation or addition.
Considering that the same event argument span could play different role in different event types~\cite{sheng-etal-2021-casee}, the decision of event arguments are conditioned on the event type.
Hence, $\phi$ should imply the conditional dependency between event types and tokens.
Accordingly, we adopt  Conditional Layer Normalization (CLN)~\cite{Su19:CLN} to implement $\phi$.
CLN is mostly based on the Layer Normalization~\cite{Ba16:LN}, but it dynamically computes the gain $\gamma$ and bias $\beta$ based on the prior condition instead of directly deploying them as learnable parameters in neural networks.
Given the event type representation $\mathbf{e}_j$ as condition and a token representation $\mathbf{h}_i$, the fusion function $\phi$ is achieved via CLN as:
\begin{equation}
\begin{aligned}
& \phi(\mathbf{e}_j, \mathbf{h}_i)  = \mathtt{CLN}(\mathbf{e}_j, \mathbf{h}_i) = \gamma_j \odot (\frac{\mathbf{h}_i - \mu_i}{\sigma_i}) + \beta_j , \\
& \gamma_j = \mathbf{W}_{\gamma} \mathbf{e}_j + \mathbf{b}_{\gamma} , \beta_j = \mathbf{W}_{\beta} \mathbf{e}_j + \mathbf{b}_{\beta},
 \end{aligned}
\end{equation}
where $\mu_i \in \mathbb{R}$ and $\sigma_i \in \mathbb{R}$ are the mean and standard variance taken across the elements of $\mathbf{h}_i$, and $\gamma_j \in \mathbb{R}^{d}$ and $\beta_j \in \mathbb{R}^{d}$ are respectively the conditional gain and bias.
In this way, the event type information is expressed as conditional information, and is thus integrated with token representations.

\subsubsection{Grid Representation}
We employ two semantic fusion functions, $\phi^c, \phi^r$, to respectively derive entry representations for the cause and effect grid table .
Each semantic fusion function is implemented by a layer of CLN, and thus the entry representation is obtained as: 
\begin{equation}
\begin{aligned}
& \mathbf{g}_{j,i}^c = \phi^c(\mathbf{e}_j, \mathbf{h}_i) =  \mathtt{CLN}^c(\mathbf{e}_j, \mathbf{h}_i), \\
& \mathbf{g}_{j,i}^r = \phi^r(\mathbf{e}_j, \mathbf{h}_i) =  \mathtt{CLN}^r(\mathbf{e}_j, \mathbf{h}_i),
\end{aligned}
\end{equation}
where $\mathbf{g}_{j,i}^c$, $\mathbf{g}_{j,i}^r$ are respectively the entry representation in the cause and effect table for grid tagging.

\subsection{Training and Inference}
Since multiple tags could be simultaneously assigned towards $(e_j, t_i)$ in each table, we conduct multi-label classification upon entry representations.
Specifically, a fully-connected network  predicts the  probability of each tag for $(e_j, t_i)$ as:
\begin{equation}
\mathbf{p}^{\mathcal{I}}_{j, i} =  \mathtt{sigmoid}(\mathbf{g}_{j,i}^{\mathcal{I}}\mathbf{W}^{\mathcal{I}} + \mathbf{b}^{\mathcal{I}} )
\end{equation}
where $\mathcal{I} \in \{c, r\}$ is  the symbol of grid field denoting the cause and effect grid table respectively, and each dimension of $\mathbf{p}^{\mathcal{I}}_{j, i}$ denotes the probability for a tag between $(e_j, t_i)$. 
Consequently, we  adapt Cross-Entropy loss as the loss function:
\begin{equation}
\label{equ:loss1}
\mathcal{L}^{\mathcal{I}} = -\sum_{j=1}^{m}\sum_{i=1}^{n}\sum_{k \in \mathit{C}  } \mathbb{I}(y^{\mathcal{I}}_{ji}=k) \mathtt{log}(\mathbf{p}^{\mathcal{I}}_{j,i}[k]), 
\end{equation}
where $\mathit{C}$ is the set of predefined tags, $\mathbf{p}^{\mathcal{I}}_{j,i}[k] \in [0, 1]$ is the predicted probability of tag $k$ between $(e_j, t_i)$ and $y^{\mathcal{I}}_{ji}$ is the ground truth tag between  $(e_j, t_i)$.
$\mathbb{I}$ is a switching function which equals to $1$ when $y^{\mathcal{I}}_{ji}=k$, otherwise 0.
Following equation~\ref{equ:loss1}, we obtain losses from both grid tables, and aggregate them for the final training objective:
\begin{equation}
\label{equ:loss}
\mathcal{J}(\theta) = \mathcal{L}^{c} + \mathcal{L}^{r}.
\end{equation}

For inference, $\mathbf{p}^{\mathcal{I}}_{j, i}$ is converted into tags whose probability overweights the scalar threshold  $\tau^{\mathcal{I}} \in [0, 1]$, which is a  manually tuned hyper-parameter.

\section{Experiments}

\subsection{Dataset and Evaluation}
\textbf{Dataset} We conduct experiments on the corpus~\cite{tianchi2021EventCausal} released by China Conference on Knowledge Graph and Semantic Computing 2021 (CCKS2021).
The corpus comes from the public news and reports, containing 7,000 sentences with an average length of 104 tokens.
It annotates 15,816 events containing 7908 event causality pairs, covering 39 types of events and 3 types of event roles, namely \texttt{Product}, \texttt{Region} and \texttt{Industry}.
To adapt this corpus into ECE task, we divide the corpus into training/validation/test set based on Cause-Effect event types. 
Specifically, CCKS2021 is divided into training/validation/test set with the proportion of 8 : 1 : 1.
We rename the split dataset as ECE-CCKS.

\noindent \textbf{Evaluation} We evaluate our model using Precision (P), Recall (R) and Micro-F1 (F1) of three metrics.
(1) \textbf{Event Argument Extraction (EAE) Metric} evaluates the model's ability to extract event arguments of interests.
Like prior works ~\cite{yang-etal-2019-exploring}, an argument is correctly predicted when its event type, span and event role  simultaneously match  the gold label.
(2) \textbf{Cause-Effect Type (CET) Metric} measures  whether both the predicted cause and effect event type match the golden answer.
(3) \textbf{Event Causality Extraction (ECE) Metric }  synthesizes the above two metrics, where an argument in ECE is  correctly extracted when its predicted cause-effect event type, span and event role simultaneously meet the gold label.

\subsection{Implementation Details}
We employ BERT$_{\rm{base}}$~\cite{devlin-etal-2019-bert} as the  encoder for our model and baselines.
For DualCor, we manually tune all the hyper-parameters on the validation set.
AdamW with learning rate of 3e-5 is adopted for model optimization.
The model is trained 10 epoches with batch size of 8.
The max length of sentence is 150 by padding shorter sentences and cutting longer ones.
The threshold $\tau^{c}, \tau^{r}$ are both set as 0.5.

\subsection{Baselines}

We employ a variety of baselines which could be classified into two streams. 

\noindent \textbf{Event-then-Causality methods}. These methods first extract events from texts and then classify the causal relation.
For event extraction, we choose three typical models.
(1)\textbf{BERT-Softmax}~\cite{devlin-etal-2019-bert} adopts BERT to learn textual representations, and conducts sequence labeling for event extraction; 
(2) \textbf{BERT-CRF} utilizes conditional random field (CRF) to capture label dependencies upon the textual representations~\cite{du-cardie-2020-document}.
(3) \textbf{DMBERT}~\cite{wang-etal-2019-adversarial-training} adopts dynamic multi-pooling~\cite{chen-etal-2015-event} upon BERT to aggregate features for event extraction.
(4) \textbf{PLMEE}~\cite{yang-etal-2019-exploring} further adopts role-specific argument tagger upon BERT to solve the argument overlapping issue. 
After obtaining the events, we enumerate all possible cause-effect pairs and follow ~\citeauthor{zuo-etal-2021-learnda}~\shortcite{zuo-etal-2021-learnda} to build a Multilayer Perceptron classifier to decide the causality. 

\noindent \textbf{Event-with-Causality methods}.
Instead of separately deriving events and causality, these methods conduct event extraction with the causality taken into consideration and thus simultaneously derive the events and causality pair.
To do this, we adapt three typical RTE methods as follows.
(1) \textbf{Novel-tagging} designs a unified label space combining causality component (cause/effect), event types, event roles and argument boundary, and conducts ECE via sequence-labeling following ~\citeauthor{zheng-etal-2017-joint}~\shortcite{zheng-etal-2017-joint}. 
(2) \textbf{CasECE}, which is inspired by CasRel~\cite{wei-etal-2020-novel}, first extracts the cause event,  conditioned on which to derive the effect event. 
(3) \textbf{Pair-linking} works in a grid tagging manner following ~\citeauthor{wang-etal-2020-tplinker}~\shortcite{ wang-etal-2020-tplinker}. It first conducts  event-type-level pair linking to derive the cause-effect event-type, which is then used as conditional information for token-pair linking to derive event arguments.
Appendix~\ref{appendix:baselines} provides details about how we adapt these methods for ECE.

\begin{table*}[t]
	\centering\footnotesize\setlength{\tabcolsep}{5pt}
	\begin{tabular*}{1 \textwidth}{@{\extracolsep{\fill}}@{}lccccccccc@{}}
		\toprule
		\multicolumn{1}{c}{} & \multicolumn{3}{c}{EAE(\%)} & \multicolumn{3}{c}{CET(\%)} & \multicolumn{3}{c}{ECE(\%)} \\
		\cmidrule{2-4}
		\cmidrule{5-7}
		\cmidrule{8-10}
		\multicolumn{1}{c}{} & P & R & F1 & P & R & F1 & P & R & F1  \\
		\midrule
		BERT-softmax+Causality &   32.55 & 35.11 & 33.78  &  49.61 & 64.20 & 55.97  & 30.47 & 31.52 &  30.99 \\ 
		BERT-CRF+Causality  & 35.52 & 34.10 &  34.79  & 53.22 & 60.95 & 56.82  & 31.02 &  31.28 &  31.15 \\
		DMBERT+Causality &   34.27 & 38.18 & 36.12 &  52.87 & \bf 63.20 &  57.58 & 30.08 &  34.93 &  32.33 \\ 
		PLMEE+Causality & 34.22 &  40.70 &  37.18 &  58.11 & 60.20 & 59.13 & 29.98 & 41.14 & 34.69 \\
		\midrule
		Novel-tagging  & \bf 59.40 &  28.47 &  38.49  &  49.79 &  61.70 &  55.11 &  51.52 & 26.75 & 35.22  \\ 
		CasECE &   36.88 &  36.72 &  36.80 &  58.26 &  59.70 &  58.97  &  31.30 &  41.81 &   35.80 \\ 
		Pair-tagging &  47.08 & 46.49 &   46.79 &   55.78 &  \bf 62.95 & 59.14 &   39.24 &   \bf 47.69 &  43.05 \\ 
		\midrule
		 \bf DualCor & 58.05 & \bf 47.60 & \bf 52.31 &  \bf 61.75 &  58.19 & \bf 59.92 & \bf 48.56 & 44.85 &  \bf 46.63 \\
		\bottomrule
	\end{tabular*}
	\caption{Overall results. The Wilcoxons test shows significant difference (p$<$0.05) between DualCor and baselines.}
	\label{tab:performance}
\end{table*}

\subsection{Main Results}
We report the overall results in   Table~\ref{tab:performance}, and have observations as follows.

(1) The event-then-causality baselines generally produce weak performances, especially on the Precision indicator.
The reason lies in that these methods extract events without considering the interest of causality.
As a result, many causality-unrelated events are wrongly extracted, which would confuse the causality decision.

(2) Performances of the event-with-causality baselines are superior to the event-then-causality models, since the events are extracted with causality modeling, thus reducing the number of redundant events.
However, their performances are still barely satisfactory, since the entity-oriented relation modeling strategy could not sufficiently to explore intra- and inter- correlations between events.

(3) DualCor achieves the best results among all baselines, we attribute this to that our designed dual grid tagging schema effectively explore the intra- and inter-event argument correlations. 
Despite of this, the overall ECE performance is far from satisfactory. 
This reflects that ECE requires investigations from future works to improve it.

\subsection{Single pair vs. Multi pairs}

\begin{figure}
	\centering
	\includegraphics[width=0.85\linewidth]{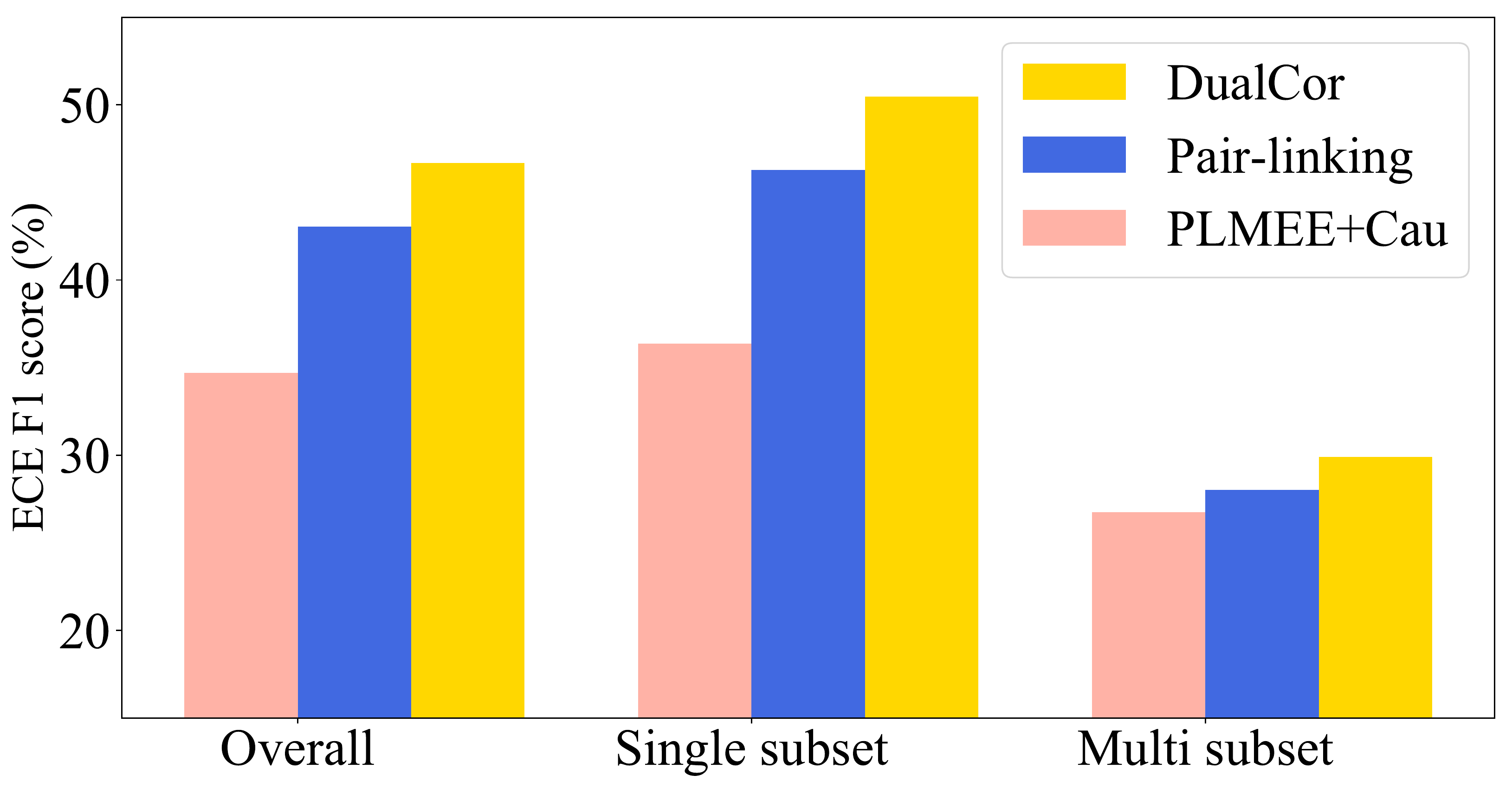}
	\caption{ECE performances on overall test set, Single and Multi subset. Appendix~\ref{appendix:single-multi} shows detailed values.}
	\label{fig:multi}
\end{figure}

We notice that nearly 10\% sentences in our dataset express multiple event causality pairs, and thus probe how the number of causality pairs influences the ECE performance.
Specifically, we divide the test set into a \textbf{Single} subset where each sentence contains only one event causality pair, otherwise,  \textbf{Multi} subset.
Apart from DualCor, PLMEE+Causality (PLMEE+Cau in short) and Pair-linking are chosen as representatives for compared baselines, and we present their performances in   Figure~\ref{fig:multi}. 
Reading from the figure, we could see that (1)  all models present a decreasing performance from Single to Multi subset, reflecting that ECE towards multiple causality pairs is much tricky.
(2) Reasons for the weak performance on the Multi subset may be that the increasing number of causality pairs come from the increase of mentioned events, which demands more complicated inter-event correlations modeling~\cite{DBLP:conf/sigir/ShengSGCCWLX22}. 
(3) Since the performance on the Multi subset is obviously inferior to the overall and Single subset performances, we argue that \textbf{Multi-pairs}  could be one great challenge which deserves investigation from future ECE works.

\section{Analysis and Discussion}
\subsection{Ablation Study}
\begin{table}[t]
	\centering\footnotesize\setlength{\tabcolsep}{2.5pt}
	\begin{tabular*}{0.48 \textwidth}{@{\extracolsep{\fill}}@{}lcccc@{}}
		\toprule
		\bf Method & EAE & CET  & ECE  \\
		\midrule
		\bf DualCor  & \bf 52.50 & \bf 61.60 &  \bf 47.58 \\
		\midrule
		w/o Intra Cor &  20.47 &  14.38 &  10.37  \\
		w/o Inter Cor & 48.57 &  56.82 &  43.36 \\
		\midrule
		w/o type-aware encoding & 47.69 & 56.00 &  43.16 \\
		$\phi \rightarrow$  Concatenation &  51.39 & 59.56 &  45.88   \\
		$\phi  \rightarrow $ Addition & 51.96 & 61.08 & 46.83  \\
		\bottomrule
	\end{tabular*}
	\caption{Ablation Study (F1\%) on the validation set . Appendix~\ref{appendix:test-ablation} illustrates ablation on the test set.}
	\label{tab:ablation}
\end{table}
To study how each module contributes to the performance, we ablate to DualCor on the validation set as Table~\ref{tab:ablation} shows. 

\begin{table*}[t]
	\centering\footnotesize\setlength{\tabcolsep}{5pt}
	\begin{tabular*}{0.93 \textwidth}{@{\extracolsep}@{}cc}
		\toprule
		\textbf{Category} & \textbf{Example} \\
		\midrule
		\makecell[c]{\bf Wrong Cause-\\\bf Effect Type} & \makecell[l]{ \underline{Instance\#1:} The falling of stainless steel prices was caused by the drop in the cost  of pure nickel. \\ Gold: \{Event type$_\mathtt{cause}$: \textcolor{magenta}{Cost Declination}, event type$_\mathtt{effect}$: \textcolor{magenta}{Price Declination} \} \\ Predicted: \{Event type$_\mathtt{cause}$: \textcolor{blue}{Price Declination}, event type$_\mathtt{effect}$: \textcolor{magenta}{Price Declination} \} } \\
		\midrule
		\makecell[c]{ \bf Redundant \\ \bf Arguments} & \makecell[l]{ \underline{Instance\#2:} Feed prices rise across the country, reducing  the profits in poultry industry. \\ Gold: \{Region$_\mathtt{cause}$: \textcolor[RGB]{218, 165, 32}{across the country}, Region$_\mathtt{effect}$: \textcolor[RGB]{218, 165, 32}{None} \}  \\ Predicted: \{Region$_\mathtt{cause}$: \textcolor[RGB]{218, 165, 32}{across the country}, Region$_\mathtt{effect}$: \textcolor[RGB]{34, 139, 34}{across the country} \}  } \\
		\midrule
		\makecell[c]{ \bf Missing \\ \bf Arguments} & \makecell[l]{\underline{Instance\#3:} 50\% of coke enterprises in Shanxi, Ningxia and 30\% of those  in Inner Mongolia have  \\ restricted their production, for which the coke output decreased.\\ Gold: \{Region$_\mathtt{reason}$: \textcolor[RGB]{178,34,34}{Shanxi, Ningxia, \textit{Inner Mongolia}} \}  \\ Predicted: \{Region$_\mathtt{reason}$: \textcolor[RGB]{178,34,34}{Shanxi, Ningxia} \} } \\
		\bottomrule
	\end{tabular*}
	\caption{Error analysis, where we only present the associated event types and event arguments due to the space limitation. Appendix~\ref{appendix:error} provides the complete event causality pair for these three instances. }
	\label{tab:error_ana}
\end{table*}

\begin{table}
	\centering\footnotesize\setlength{\tabcolsep}{7pt}
	\begin{tabular}{crrr}
		\toprule
		\bf  & \bf \#Para.  & \bf Training & \bf Inference   \\ 
		\midrule
		DualCor    & 107.10M & 18.6sents/s   & 38.9sents/s  \\
		Pair-linking & 107.63M   & 6.4sents/s   & 19.4sents/s   \\
		\bottomrule
	\end{tabular}
	\caption{Efficiency comparison.}
	\label{tab:efficiency}
\end{table}

We probe the argument correlations via ablation to the tagging scheme.
(1) w/o Intra-event argument correlations (Intra-Cor): To explore the necessity of Intra-Cor, we remove  tags  whose correlation-field are $\mathtt{Intra}$ in the tagging scheme.
This leads to the sharp performance drops since Intra-Cor is the key to derive individual event from each grid.
(2) w/o Inter-event argument correlations (Inter-Cor): To certify the effectiveness of Inter-Cor, we remove tags whose correlation-field are $\mathtt{Inter}$.
Without Inter-cor, the causality pairs are obtained by exhaustive enumeration between the cause and effect event which are individually derived from two tables.
The ECE performance declines 4.42\%, reflecting the importance of Inter-Cor.
(3) We observe that the removing of either type of tags would hurt performances, verifying that these two correlations are both beneficial and functional for ECE.

We explore the influence of  the model architecture via ablation to the  encoding and grid representation layer.
(1) w/o type-aware encoding: Instead of the collaborative encoding as Equation~\ref{equ:encoding} shows, when we encode sentence using BERT while obtain event type embeddings by random initialization, the final performance declines by 4.42\%.
This manifests the importance of capturing semantic dependency between event types and each sentence.
(2) $\phi \rightarrow$  Concatenation or Addition: To explore the impact of the semantic fusion function $\phi$ in Section~\ref{sec:fusion}, we respectively replace CLN  as concatenation and addition.
The performance degradation upon two variants signifies that CLN could better enhance token representations with event types, producing more expressive entry representations.

\subsection{Efficiency Discussion} 

Since Pair-Linking also works in a grid tagging manner and  achieves the comparable performance with DualCor, we discuss the efficiency of these two architectures from two aspects: parameter amount and running speed.
For the sake of fairness, we run them on the same GPU server.
Reading from Table~\ref{tab:efficiency}, we notice that the amount of parameters of our model and Pair-Linking is roughly equal.
We attribute this to that they both   exploits the same basic encoding and grid representations learning strategy.
However, we observe that the training and inference speeds of our model are respectively about 2.91 and 2.01 times faster than Pair-Linking.
This is mainly because that the representation learning for two grids  are carried in parallel in our model, while those of Pair-Linking are sequentially conducted.
Considering analysis above, we could conclude that our model also maintains efficiency advantage over Pair-linking.

\subsection{Error Analysis} 

To probe the drawbacks of DualCor and promote future works, we conduct error analysis towards 100  randomly selected failure instances.
Here, we discuss three typical error types as Table~\ref{tab:error_ana} shows.
(1) \textbf{Wrong Cause-Effect Type} refers to predicting the wrong combination of cause-effect event types as \underline{Instance\#1}.
This error can severely hurt the final performances, since event arguments under the wrong cause-effect type would be regarded as false positive in ECE.
We notice that almost 40\% error cases of DualCor belong to this type, while that of Pair-linking is 32\%.
We attribute this to that our method mainly focus on correlations between event arguments, which lacks exact cause-effect modeling between event types, while the  event-type-level pair linking in Pair-linking accounts for this.
(2) \textbf{Redundant Arguments}  denotes that the model predicts an argument which actually does not exist, as the redundant region for effect event in \underline{Instance\#2}.
This kind of errors usually appear  between the cause and effect event upon the same event role, which demonstrates the difficulty of deducing causality-specific event arguments.
Though redundant arguments accounts for nearly 30\% error cases of DualCor, it is almost 10\% lower than that of Pair-linking.
This reveals the importance of exploring intra- and inter- event argument correlations to discriminate the cause / effect event arguments.
(3) \textbf{Missing Arguments} refers to that the model fails to predict the existed event argument, as the missed ``Inner Mongolia'' in \underline{Instance\#3}.
We observe that it usually occurs for event roles which contains multiple event arguments, where more sophisticated modeling of intra-event arguments correlations are required.

\section{Conclusion}
In this paper, we formulate a new task, Event Causality Extraction (ECE), which aims to extract the cause-effect event pairs with structured event information from plain texts.
We propose a method based on an elaborately devised dual grid tagging scheme, which explores the intra- and inter-event argument correlations for the task.
Experiment results prove the effectiveness of our method, and extensive analyses are conducted to point out several promising directions to inspire future works.

\section*{Acknowledgements}
We would like to thank Bowen Yu and Jiangxia Cao for helpful discussions, support, and feedback on earlier versions of this work. We would also like to thank the anonymous reviewers for their insightful comments and suggestions. This work is supported by the National Key Research and Development Program of China (grant No.2021YFB3100600), the Strategic Priority Research Program of Chinese Academy of Sciences (grant No.XDC02040400) , the Youth Innovation Promotion Association of CAS (Grant No. 2021153).

\bibliography{anthology,custom}

\clearpage

\appendix

\section{Decoding strategy}
\label{appendix:decoding}

\begin{figure}[h]
	\centering
	\includegraphics[width=0.98\linewidth]{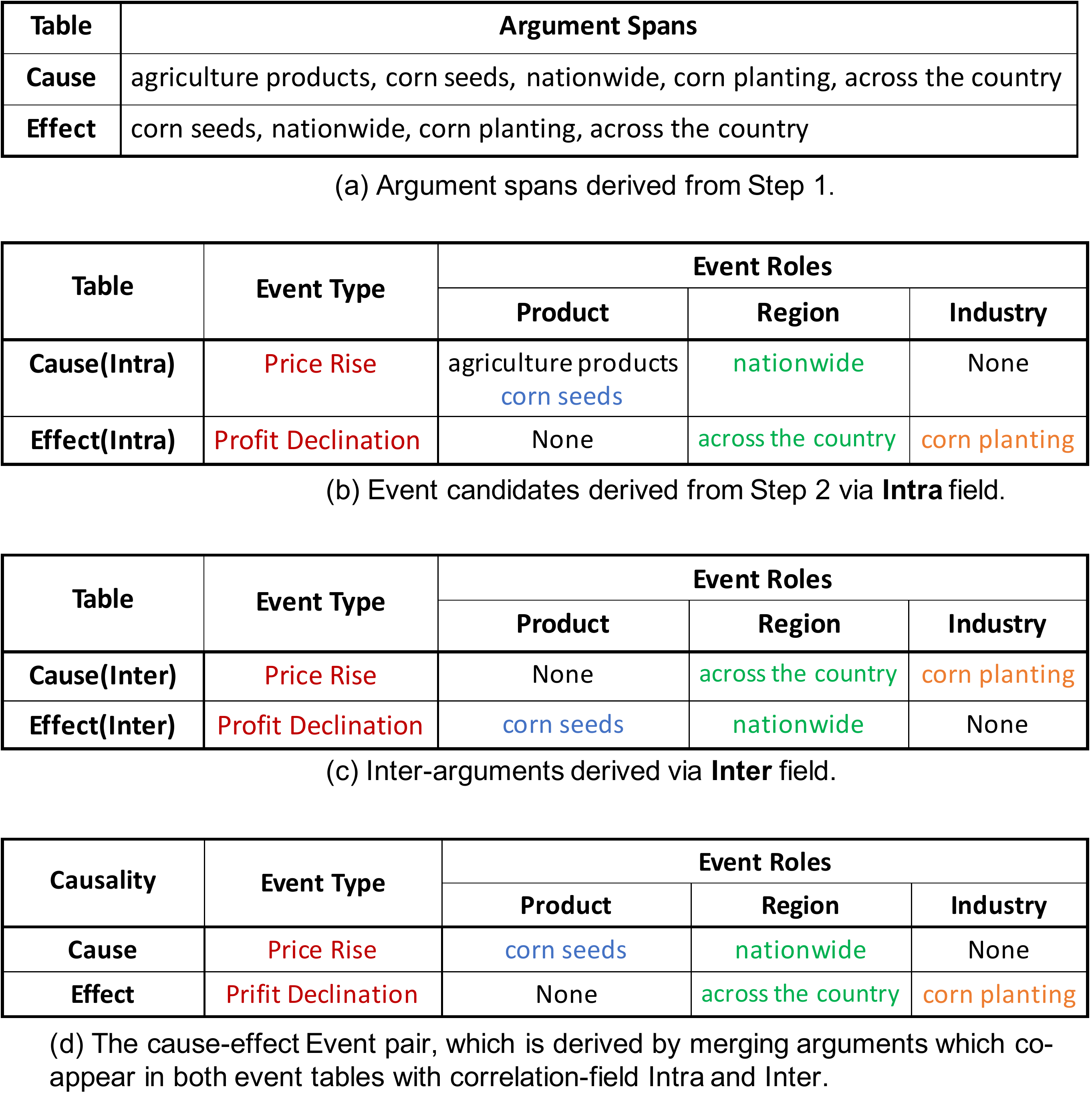}
	\caption{Detailed illustration to decoding strategy.}
	\label{fig:arguments}
\end{figure}

\textbf{Step 1. Argument span decoding.} In this stage, we derive argument spans for cause/effect events using  the nearest start-end match principle~\cite{wei-etal-2020-novel}. 
Specifically, for those entry tags having the same correlation-field and role-field in the same row, we match the start position to the nearest end position according to the position-field to obtain candidate argument spans  Figure~\ref{fig:arguments}.(a).

\textbf{Step 2. Event structure decoding.}
In this stage, we collect candidate argument spans attached to the same event types.
Specifically, we merge the event arguments with correlation-field $ \mathtt{Intra}$ belonging to the same row, resulting in structured candidate events in Figure~\ref{fig:arguments}.(b).

\textbf{Step 3. Causality pair decoding.}
To derive causality pairs, we match inter-event correlated arguments between candidate cause and effect events. 
Specifically, we first obtain event argument with correlation-field $ \mathtt{Inter}$ in each table as Figure~\ref{fig:arguments}.(c).
Then, we search the arguments co-occurring in both event tables simultaneously associating with correlation-field $\mathtt{Intra}$ and $\mathtt{Inter}$, and merge them to form the cause-effect pair in Figure~\ref{fig:arguments}.(d).

\section{Details about adapted baseliens}
\label{appendix:baselines}

We detail the adaption of RTE methods to ECE.

(1) \textbf{Novel-tagging} is adapted from ~\citeauthor{zheng-etal-2017-joint}~\shortcite{zheng-etal-2017-joint}. It performs RTE through sequence labeling with a novel tagging scheme, which combines the label spaces of relation types and relation roles (subject and object),
Similarly, we adopt a unified label space combining cause/effect, event type, event roles and argument boundary tag, namely \begin{math} \{ \mathtt{Causality\text{-}EventType\text{-}EventRole\text{-}Bdy} \} \end{math}.
Given the 2 types of causality components, 39 predefined event types, 3 predefined event roles and the Start/End boundary indicator, the capacity of the unified label space is $2 \times 39 \times 3 \times 2 = 468$. We employ the unified labels to tag the tokens in a sequence labeling manner with BERT+Softmax.
Note that we only deploy BERT+Softmax for sequence labeling  here, since the joint label space is too large for BERT-CRF to implement on our experiment devices.

(2) \textbf{CasECE} is adapted from CasRel~\cite{wei-etal-2020-novel}, which conducts RTE by modeling the relations as functions  mapping subject entity to object entity.
Similarly, we regard the causality relation as the function which maps the cause event to the effect event.
Following CasRel, we first extract the cause event, and then conditioned on it to derive the effect event.
During this process, PLMEE~\cite{yang-etal-2019-exploring} is utilized as the event extractor.

(3) \textbf{Pair-linking} is adapted from TPLinker~\cite{wang-etal-2020-tplinker}, where RTE is formulated as a token pair linking problem which aligns the boundary tokens of entity pairs under each relation type.
Similarly, we intend to respectively extract event arguments under  specific cause-effect event types.
Specifically, we first conduct event-type-level pair linking to derive the cause-effect event types.
Then, we utilize CLN to refine textual representations enhanced with cause-effect event type pair information, and conduct token-pair-linking to extract the event arguments for the specific cause and effect event.

\section{Other encoders}
\label{appendix:other-encoders}

We report  performances of DualCor using different  basic encoders in Table~\ref{tab:other-encoders}.

\section{Single pair vs. Multi pairs}
\label{appendix:single-multi}
We detail ECE performances on the overall test, Single and Multi subset in Table~\ref{tab:single-multi}.

\section{Ablation on the  test}
\label{appendix:test-ablation}

We provide ablation study on the test set in Table~\ref{tab:app-ablation}.

\section{Error analysis}
\label{appendix:error}

This section provides the complete instances for error analysis in Table~\ref{tab:app-error_complete}.
\begin{table*}[h]
	\centering\footnotesize\setlength{\tabcolsep}{5pt}
	\begin{tabular*}{1 \textwidth}{@{\extracolsep{\fill}}@{}lccccccccc@{}}
		\toprule
		\multicolumn{1}{c}{} & \multicolumn{3}{c}{EAE(\%)} & \multicolumn{3}{c}{CET(\%)} & \multicolumn{3}{c}{ECE(\%)} \\
		\cmidrule{2-4}
		\cmidrule{5-7}
		\cmidrule{8-10}
		\multicolumn{1}{c}{} & P & R & F1 & P & R & F1 & P & R & F1  \\
		\bf DualCor$_{\rm{BERT_{base}}}$  &  58.05 &  47.60 & 52.31 &   61.75 & 58.19 & 59.92 & 48.56 & 44.85 &  46.63  \\
				DualCor$_{\rm{Roberta_{base}}}$ & 61.46 & 46.29 & 52.80  &  66.14 & 58.19 &  61.91 &  52.801 & 44.89 & 48.53 \\
			DualCor$_{\rm{Roberta_{large}}}$ & 63.44 & \bf 50.48 &  56.22 & 67.52 & \bf 62.70 &  65.02 & 54.67 &  \bf 49.80 &  52.12    \\ 
				DualCor$_{\rm{MacBERT} }$ & \bf 67.64 &  49.19 & \bf 56.96 &   \bf  70.68 & 60.95 & \bf 65.45 & \bf 58.29 &   48.02 & \bf 52.66  \\ 
		\bottomrule
	\end{tabular*}
	\caption{Overall results on the test set.}
	\label{tab:other-encoders}
\end{table*}

\begin{table*}[t]
	\centering\footnotesize\setlength{\tabcolsep}{5pt}
	\begin{tabular*}{1 \textwidth}{@{\extracolsep{\fill}}@{}lccccccccc@{}}
		\toprule
		\multicolumn{1}{c}{} & \multicolumn{3}{c}{Overall(\%)} & \multicolumn{3}{c}{Single Subset(\%)} & \multicolumn{3}{c}{Multi Subset(\%)} \\
		\cmidrule{2-4}
		\cmidrule{5-7}
		\cmidrule{8-10}
		\multicolumn{1}{c}{} & P & R & F1 & P & R & F1 & P & R & F1  \\
		\midrule
		PLMEE+Causality & 29.98 & 41.14 & 34.69 & 29.89 & 46.42 & 36.37 & 30.58 &  23.76 & 26.74 \\
		Pair-linking &  39.24 &   \bf 47.69 &  43.05  &  40.31 & \bf 54.32 &  46.28 &   32.15 &  \bf 24.82 &  28.03 \\ 
		\midrule
		\bf DualCor & \bf  48.64 &  44.85 &  \bf 46.67 & \bf 49.39 &  51.56 &   \bf  50.46   & \bf 43.65 &  22.72 &  \bf 29.89  \\
		\bottomrule
	\end{tabular*}
	\caption{ECE performances on the overall test set, Single subset and Multi subset. }
	\label{tab:single-multi}
\end{table*}

\begin{table*}[t]
	\centering\footnotesize\setlength{\tabcolsep}{2.5pt}
	\begin{tabular*}{0.7 \textwidth}{@{\extracolsep{\fill}}@{}lcccc@{}}
		\toprule
		\bf Method & EAE & CET  & ECE  \\
		\midrule
		\bf DualCor  & \bf 52.36 &  \bf59.96 &  \bf 46.67  \\
		\midrule
		w/o Intra Cor &  19.11 &  12.51 &  9.32  \\
		w/o Inter Cor & 49.29 &  55.04 &  42.88 \\
		\midrule
		w/o type-aware encoding & 47.21 & 54.72  &  41.79 \\
		$\phi \rightarrow$  Concatenation &  50.91 & 57.99 &  44.04   \\
		$\phi  \rightarrow $ Addition & 51.52 & 58.90 & 45.05 \\
		\bottomrule
	\end{tabular*}
	\caption{Ablation Study: F1\% upon the three metrics on the test set.}
	\label{tab:app-ablation}
\end{table*}

\begin{table*}[t]
	\centering\footnotesize\setlength{\tabcolsep}{5pt}
	\begin{tabular*}{0.93 \textwidth}{@{\extracolsep}@{}cc}
		\toprule
		\textbf{Category} & \textbf{Example} \\
		\midrule
		\makecell[c]{\bf Wrong Cause-\\\bf Effect Type} & \makecell[l]{ \underline{Instance\#1:} The falling of stainless steel prices was caused by the drop in the cost  of pure nickel. \\ Gold: \{Event type$_\mathtt{cause}$: \textcolor{magenta}{Cost Declination}, Event type$_\mathtt{effect}$: \textcolor{magenta}{Price Declination}, \\ ~~~~~~~~~~~~Product$_\mathtt{cause}$: \textcolor{magenta}{pure nickel}, ~~~~~~~~~~~~~~ Product$_\mathtt{effect}$: \textcolor{magenta}{stainless steel}, \\ ~~~~~~~~~~~~Region$_\mathtt{cause}$: \textcolor{magenta}{None}, ~~~~~~~~~~~~~~~~~~~~~~~~~Industry$_\mathtt{effect}$: \textcolor{magenta}{None}\\ ~~~~~~~~~~~~Industry$_\mathtt{cause}$: \textcolor{magenta}{None}, ~~~~~~~~~~~~~~~~~~~~~~~Industry$_\mathtt{effect}$: \textcolor{magenta}{None}  \} \\ Predicted: \{Event type$_\mathtt{cause}$: \textcolor{blue}{Price Declination}, event type$_\mathtt{effect}$: \textcolor{magenta}{Price Declination},\\ ~~~~~~~~~~~~Product$_\mathtt{cause}$: \textcolor{magenta}{pure nickel}, ~~~~~~~~~~~~~~ Product$_\mathtt{effect}$: \textcolor{magenta}{stainless steel}, \\ ~~~~~~~~~~~~Region$_\mathtt{cause}$: \textcolor{magenta}{None}, ~~~~~~~~~~~~~~~~~~~~~~~~~Industry$_\mathtt{effect}$: \textcolor{magenta}{None}\\ ~~~~~~~~~~~~Industry$_\mathtt{cause}$: \textcolor{magenta}{None}, ~~~~~~~~~~~~~~~~~~~~~~~Industry$_\mathtt{effect}$: \textcolor{magenta}{None} \} } \\
		\midrule
		\makecell[c]{ \bf Redundant \\ \bf Arguments} & \makecell[l]{ \underline{Instance\#2:} Feed prices rise across the country, reducing  the profits in poultry industry. \\ Gold: \{ Event type$_\mathtt{cause}$: \textcolor[RGB]{218, 165, 32}{Price Rise}, ~~~~~~~Event type$_\mathtt{effect}$: \textcolor[RGB]{218, 165, 32}{Profit Declination}, \\ ~~~~~~~~~~~~~Product$_\mathtt{cause}$: \textcolor[RGB]{218, 165, 32}{feed}, ~~~~~~~~~~~~~~~~~~~~~~~~ Product$_\mathtt{effect}$: \textcolor[RGB]{218, 165, 32}{None}, \\ ~~~~~~~~~~~~ Region$_\mathtt{cause}$: \textcolor[RGB]{218, 165, 32}{across the country}, ~~~ Region$_\mathtt{effect}$: \textcolor[RGB]{218, 165, 32}{None}, \\ ~~~~~~~~~~~~~Industry$_\mathtt{cause}$: \textcolor[RGB]{218, 165, 32}{None}, ~~~~~~~~~~~~~~~~~~~~~~~Industry$_\mathtt{effect}$: \textcolor[RGB]{218, 165, 32}{poultry industry}  \}  \\ Predicted: \{ Event type$_\mathtt{cause}$: \textcolor[RGB]{218, 165, 32}{Price Rising}, ~~~~~~~Event type$_\mathtt{effect}$: \textcolor[RGB]{218, 165, 32}{Profit Declination}, \\ ~~~~~~~~~~~~~~~~~~~~Product$_\mathtt{cause}$: \textcolor[RGB]{218, 165, 32}{feed}, ~~~~~~~~~~~~~~~~~~~~~~~~~ Product$_\mathtt{effect}$: \textcolor[RGB]{218, 165, 32}{None}, \\ ~~~~~~~~~~~~~~~~~~~ Region$_\mathtt{cause}$: \textcolor[RGB]{218, 165, 32}{across the country}, ~~~~ Region$_\mathtt{effect}$: \textcolor[RGB]{34, 139, 34}{across the country}, \\ ~~~~~~~~~~~~~~~~~~~~Industry$_\mathtt{cause}$: \textcolor[RGB]{218, 165, 32}{None}, ~~~~~~~~~~~~~~~~~~~~~~~Industry$_\mathtt{effect}$: \textcolor[RGB]{218, 165, 32}{poultry industry}  \}  } \\
		\midrule
		\makecell[c]{ \bf Missing \\ \bf Arguments} & \makecell[l]{\underline{Instance\#3:} 50\% of coke enterprises in Shanxi, Ningxia and 30\% of those  in Inner Mongolia have  \\ restricted their production, for which the supply of coke output.\\ Gold: \{Event type$_\mathtt{cause}$: \textcolor[RGB]{178,34,34}{Production Restriction}, ~~~~~~~~~~~~~~Event type$_\mathtt{effect}$: \textcolor[RGB]{178,34,34}{Supply Reduction}, \\ ~~~~~~~~~~~~~Product$_\mathtt{cause}$: \textcolor[RGB]{178,34,34}{coke}, ~~~~~~~~~~~~~~~~~~~~~~~~~~~~~~~~~~~~~~~~~~~~~~ Product$_\mathtt{effect}$: \textcolor[RGB]{178,34,34}{coke}, \\ ~~~~~~~~~~~~ Region$_\mathtt{cause}$: \textcolor[RGB]{178,34,34}{Shanxi, Ningxia, \textit{Inner Mongolia}}, ~~ Region$_\mathtt{effect}$: \textcolor[RGB]{178,34,34}{None}, \\ ~~~~~~~~~~~~~Industry$_\mathtt{cause}$: \textcolor[RGB]{178,34,34}{None},  ~~~~~~~~~~~~~~~~~~~~~~~~~~~~~~~~~~~~~~~~~~~~~Industry$_\mathtt{effect}$: \textcolor[RGB]{178,34,34}{None} \} \\  Predicted: \{Event type$_\mathtt{cause}$: \textcolor[RGB]{178,34,34}{Production Restriction}, ~~~~~~~Event type$_\mathtt{effect}$: \textcolor[RGB]{178,34,34}{Supply Reduction}, \\ ~~~~~~~~~~~~~~~~~~~~Product$_\mathtt{cause}$: \textcolor[RGB]{178,34,34}{coke}, ~~~~~~~~~~~~~~~~~~~~~~~~~~~~~~~~~~~~~~~ Product$_\mathtt{effect}$: \textcolor[RGB]{178,34,34}{coke}, \\ ~~~~~~~~~~~~~~~~~~~ Region$_\mathtt{cause}$: \textcolor[RGB]{178,34,34}{Shanxi, Ningxia}, ~~~~~~~~~~~~~~~~~~~~~~ Region$_\mathtt{effect}$: \textcolor[RGB]{178,34,34}{None}, \\ ~~~~~~~~~~~~~~~~~~~~Industry$_\mathtt{cause}$: \textcolor[RGB]{178,34,34}{None},  ~~~~~~~~~~~~~~~~~~~~~~~~~~~~~~~~~~~~~~Industry$_\mathtt{effect}$: \textcolor[RGB]{178,34,34}{None}  \} } \\
		\bottomrule
	\end{tabular*}
	\caption{The complete results of error analysis. }
	\label{tab:app-error_complete}
\end{table*}

\end{document}